\documentclass{elsarticle}

\usepackage{lineno,hyperref}
\usepackage{epstopdf}
\usepackage{bm}
\usepackage{makecell}
\modulolinenumbers[5]
\usepackage{mathrsfs}
\usepackage{amsfonts}
\usepackage{multirow}

\journal{Journal of \LaTeX\ Templates}

\begin{document}

\begin{frontmatter}

\title{Supervision-by-Hallucination-and-Transfer: A Weakly-Supervised Approach for Robust and Precise Facial Landmark Detection \tnoteref{mytitlenote}}

\author[firstauthor,firstauthor2]{Jun Wan}
\author[firstauthor]{Yuanzhi Yao}
\author[firstauthor2]{ Zhihui Lai\corref{cor1}}
\author[firstauthor2]{Jie Zhou}
\author[firstauthor2]{Xianxu Hou}
\author[thirdauthor]{Wenwen Min}

\cortext[cor1]{Corresponding author: lai\_zhi\_hui@163.com (Zhihui Lai).  \\ \indent \hspace{1em}}

\address[firstauthor]{School of Information Engineering, Zhongnan University of Economics and Law, Wuhan 430073, China}
\address[firstauthor2]{College of Computer Science and Software Engineering, Shenzhen University, \\ Shenzhen, Guangdong 518060, China}
\address[thirdauthor]{School of Information Science and Engineering,
	Yunnan University, Kunming, Yunnan 650091, China}
\begin{abstract}
	
High-precision facial landmark detection (FLD) relies on high-resolution deep feature representations. However, low-resolution face images or the compression (via pooling or strided convolution) of originally high-resolution images hinder the learning of such features, thereby reducing FLD accuracy. Moreover, insufficient training data and imprecise annotations further degrade performance. To address these challenges, we propose a weakly-supervised framework called Supervision-by-Hallucination-and-Transfer (SHT) for more robust and precise FLD. SHT contains two novel mutually enhanced modules: Dual Hallucination Learning Network (DHLN) and Facial Pose Transfer Network (FPTN). By incorporating FLD and face hallucination tasks, DHLN is able to learn high-resolution representations with low-resolution inputs for recovering both facial structures and local details and generating more effective landmark heatmaps. Then, by transforming faces from one pose to another, FPTN can further improve landmark heatmaps and faces hallucinated by DHLN for detecting more accurate landmarks. To the best of our knowledge, this is the first study to explore weakly-supervised FLD by integrating face hallucination and facial pose transfer tasks. Experimental results of both face hallucination and FLD demonstrate that our method surpasses state-of-the-art techniques.

\end{abstract}


\end{frontmatter}


\section{Introduction}
\label{}

Face alignment, also known as facial landmark detection (FLD), focuses on predicting facial landmarks with specific semantics. The performance of FLD directly influences various downstream tasks such as face recognition \cite{cao2016pose, Zhao2023FM3DFRFM, zhao2019novel}, face animation \cite{Peng2023EmoTalkSE}, and facial expression recognition \cite{ma2022multi, shahid2023squeezexpnet, Liu2025FacialER}. As a result, it has garnered extensive attention.

\begin{figure}[t]
\begin{center}
	\includegraphics[width=0.88\linewidth]{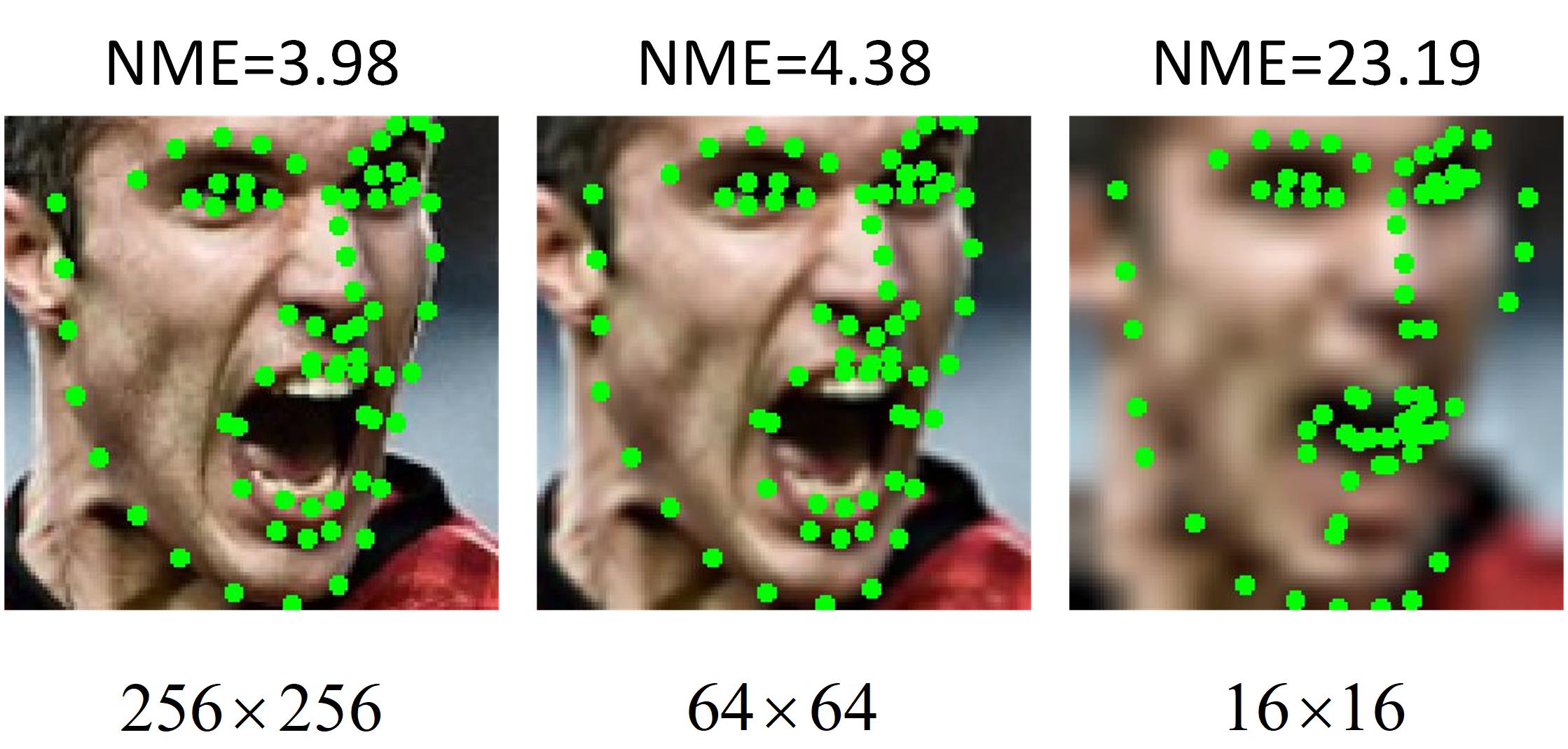}
\end{center}
\caption{Comparison of facial landmark detection results for images with different resolutions. The facial structure (e.g. face contour) of low-resolution face images (e.g. 16x16) is destroyed, which results in imprecise landmark detection (i.e., NME=23.19). While by taking low-resolution face images as inputs, our proposed SHT can learn high-resolution representations, which helps recover/preserve good facial structure and local details for precise low-resolution facial landmark detection.}
\label{fig1}
\end{figure}

Deep learning-based \cite{Wan2025FineGrainedIC, Wan2025InterpretableFL} facial landmark detection (FLD) methods have made significant progress, with high-resolution deep features playing a critical role in achieving strong performance \cite{Chandran2020AttentionDrivenCF, Lin2021StructureCoherentDF, Wan2023PreciseFL, Wan2024PreciseFL, Hu2025ProtoFormerUF}. Typically, these high-resolution features depend on high-resolution input. However, obtaining high-resolution face images in real-world monitoring scenarios remains challenging. The loss or incompleteness of facial structure and texture information in low-resolution images (as shown in Fig .\ref{fig1}) degrades detection accuracy. Moreover, the performance of FLD is also limited to insufficient labeled high-resolution training samples. Although it is easy to obtain constrained and informative high-resolution face images or videos, annotating them requires considerable human and financial resources. Fortunately, face hallucination (i.e., face super-resolution) \cite{Cao2017AttentionAwareFH, Chen2018FSRNetEL, Ma2020DeepFS} can learn high-resolution representations by learning a mapping from low-resolution images to high-resolution ones and the weakly-supervised or semi-supervised techniques \cite{Zhou2017Towards3H, Chen2019WeaklySupervisedDO, Kocabas2019SelfSupervisedLO} have also proven effective in utilizing unlabeled data to enhance deep learning tasks. However, how to achieve more precise low-resolution FLD by incorporating face hallucination tasks with unlabeled high-resolution face images or videos is still an unsolved problem.

\begin{figure*}
\begin{center}
	\includegraphics[width=0.98\linewidth]{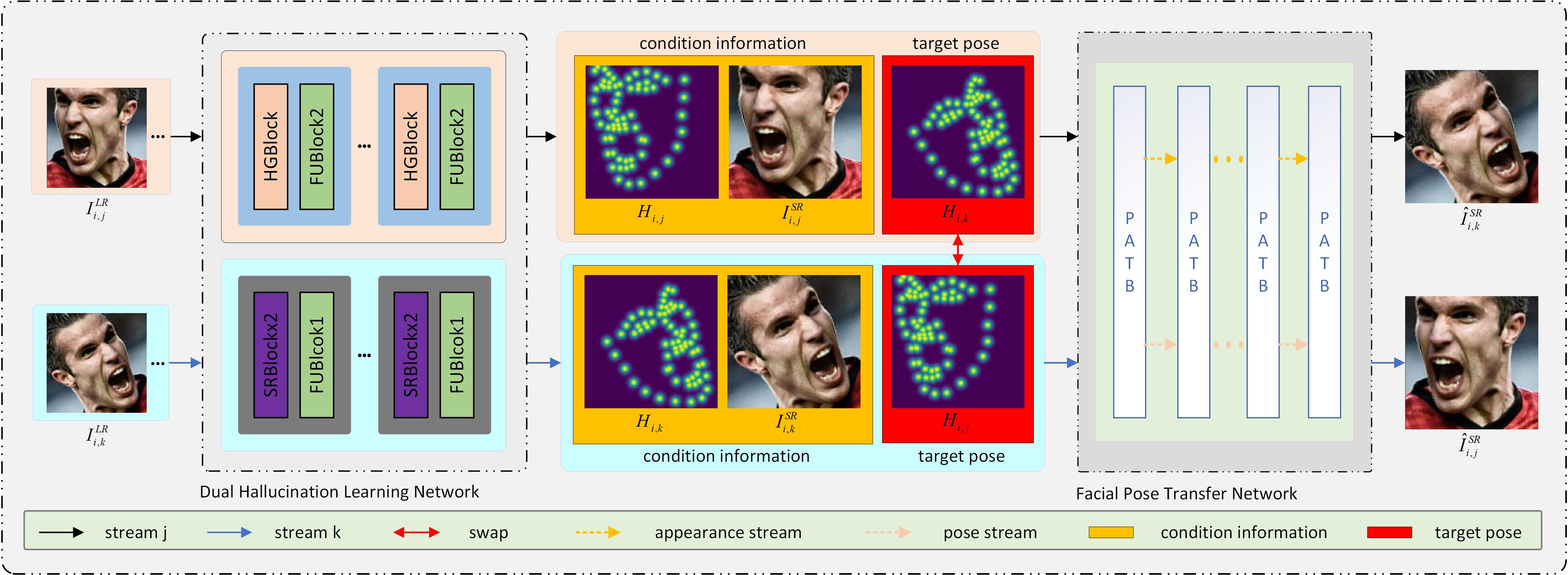}
\end{center}
\caption{The overall architecture of the proposed SHT. SHT contains two modules: Dual Hallucination Learning Network (DHLN) and Facial Pose Transfer Network (FPTN). By seamlessly integrating the DHLN and FPTN in a novel weakly-supervised framework, our SHT can simultaneously achieve more effective facial landmark detection and face hallucination.}
\label{fig2}
\end{figure*}

To tackle the aforementioned challenges, we propose a novel weakly-supervised framework called Supervision-by-Hallucination-and-Transfer (SHT) (as illustrated in Fig. \ref{fig2}), which seamlessly integrates facial landmark detection, face hallucination, and facial pose transfer tasks. Specifically, SHT consists of two key modules: the Dual Hallucination Learning Network (DHLN) and the Facial Pose Transfer Network (FPTN). The proposed DHLN consists of a landmark heatmap hallucination stream and a face hallucination stream and its creation is that we embed face hallucination task into the existing facial landmark detection pipelines, thus two streams of DHLN can be boosted by each other for hallucinating faces without distortions and generating more accurate landmark heatmaps. After that, FPTN is designed to transform the hallucinated face from one pose (i.e., the hallucinated landmark heatmaps) to another for further refining DHLN. Notably, FPTN is not required during inference, resulting in no additional computational overhead. Finally, by integrating landmark heatmap hallucination stream, face hallucination stream and FTPN, the proposed SHT can use unlabeled high-resolution face images or videos (more facial prior knowledge) to improve both facial landmark detection and face hallucination tasks. We summarize the contributions as follows:

1) By incorporating the landmark heatmap hallucination and the face hallucination, a novel DHLN is proposed, in which high-resolution representations can be gradually learned from low-resolution inputs for helping recover/preserve good facial structure and local details and produce more effective landmark heatmaps. 

2) A well-designed FPTN is presented to further enhance both the landmark heatmap hallucination and face hallucination tasks. By treating these tasks as intermediate supervision signals and utilizing unlabeled high-resolution face images or videos, FPTN leverages additional facial prior knowledge to improve the accuracy of low-resolution FLD. Importantly, FPTN does not run during inference, ensuring no additional computational overhead.

3) A novel framework called SHT is developed to incorporate DHLN and FPTN for low-resolution FLD and facial hallucination problems. To the best of our knowledge, this is the first study to address precise low-resolution FLD by incorporating face hallucination and facial pose transfer tasks in a weakly-supervised framework. Experimental results demonstrate that our method delivers outstanding performance in both face hallucination and facial landmark detection.

In the next section, we review related work on facial landmark detection and facial hallucination. Then, the proposed DHLN, FPTN and SHT will be presented in Section III. Section IV describes the corresponding experiments and Section V concludes the paper.

\section{Related Work}
This section reviews related works in FLD and face hallucination.

\textbf{Facial landmark detection. }Early FLD methods \cite{cootes1995active, cootes2001active, Cristinacce2006FeatureDA} as well as their variations \cite{Tzimiropoulos2014GaussNewtonDP, Liu2010VideobasedFM} improve the performance by adjusting the facial shape and appearance model parameters and they remain sensitive to variations of facial poses and occlusions. Then, two kinds of methods emerge in the FLD field. The first one is coordinate regression-based methods \cite{Wan2018FaceAB, wan2018face, Xia2022SparseLP, Li2022TowardsAF}, which directly regresses landmark coordinates using full connection layers, thus leading to the loss of spatial information and reducing the detection accuracy. Therefore, the second kind of method (i.e., heatmap regression-based methods \cite{Dong2018StyleAN, Liu2019SemanticAF, Kumar2020LUVLiFA, Chandran2020AttentionDrivenCF, Wan2020RobustFL, Wan2021RobustAP, huang2021adnet, McCouat2022ContourHuggingHF}) has been proposed by producing a heatmap for each facial landmark. The heatmap regression-based method has become the mainstream method in FLD and has achieved remarkable performance. However, they also experience significant performance degradation when applied to low-resolution inputs, limiting their usability in certain practical scenarios.

The proposed SHT method addresses these issues from two aspects: 1) by integrating FLD with face hallucination tasks, it learns high-resolution representations from low-resolution inputs, thereby recovering and preserving crucial facial structures and local details; 2) these high-resolution representations enable more precise heatmap regression, leading to improved accuracy in landmark detection.

\begin{figure*}[t]
\begin{center}
	\includegraphics[width=0.98\linewidth]{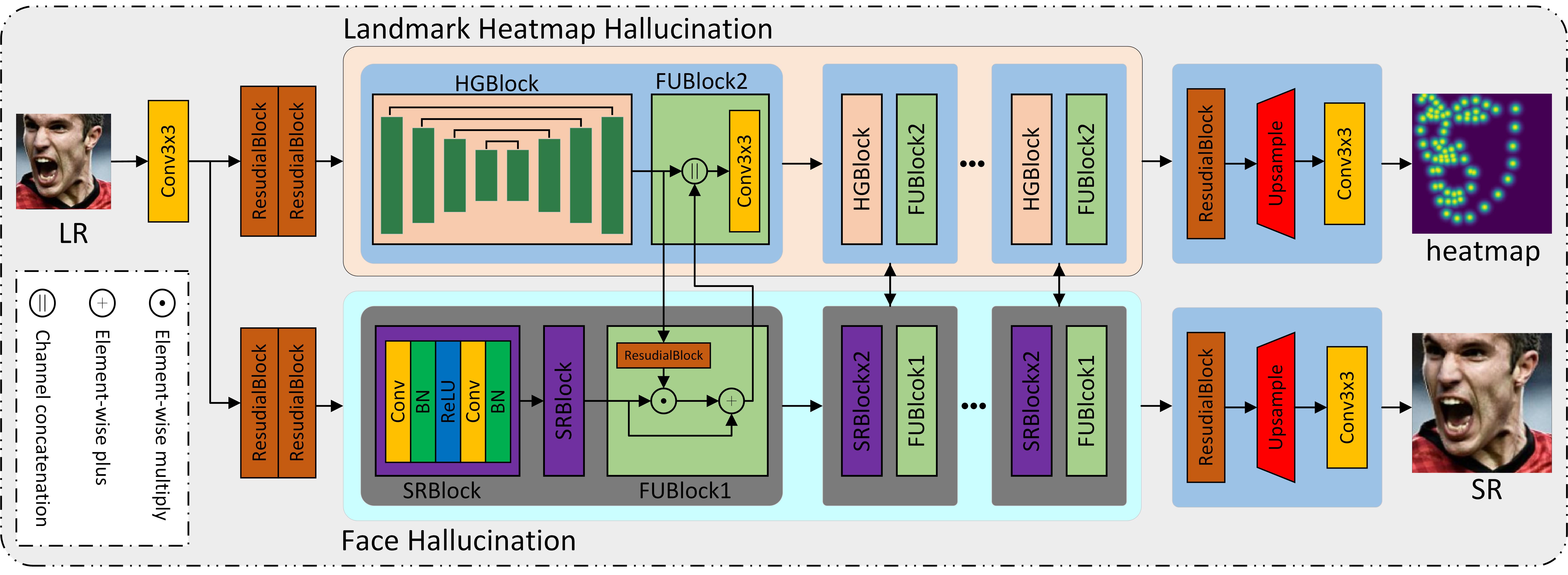}
\end{center}
\caption{The overall architecture of the proposed DHLN. By deep collaboration of landmark heatmap hallucination stream and face hallucination stream, DHLN can learn high-resolution representations with low-resolution input to recover/preserve good facial structure and local details, which help achieve more accurate FLD. Meanwhile, facial structure information learned from the landmark heatmap hallucination stream can also be applied to boost the face hallucination stream.}
\label{fig3}
\end{figure*}

\textbf{Face hallucination.} Current face hallucination methods \cite{Yu2016UltraResolvingFI, Huang2017WaveletSRNetAW, Zhang2018SuperIdentityCN, Cao2017AttentionAwareFH, Yu2018FaceSG, Ma2020DeepFS} either focus on hallucinating the entire face at once or synthesizing the whole face by hallucinating local parts. For instance,  Yu et al. \cite{Yu2016UltraResolvingFI} use a deep discriminative generative network for face hallucination tasks, while Attention-FH \cite{Cao2017AttentionAwareFH} leverages the relationships between different facial parts through a Markov decision process to generate more realistic faces. Additionally, several methods \cite{Yu2018FaceSG, Chen2018FSRNetEL, Bulat2018SuperFANIF, Ma2020DeepFS} try to boost face hallucination by exploiting facial prior knowledge. FSRNet \cite{Chen2018FSRNetEL}, for example, can hallucinate very low-resolution face images without the requirement of good alignment by making full use of facial landmark heatmaps and parsing maps as the geometry prior. Super-FAN \cite{Bulat2018SuperFANIF} utilizes GANs to hallucinate the low-resolution images and then incorporates facial structural information by incorporating the face alignment task. Ma et al. \cite{Ma2020DeepFS} propose a deep face hallucination method that iteratively collaborates with the face alignment task to better utilize facial priors. More recently, Bao et al. \cite{Bao2022AttentionDrivenGN} introduce AD-GNN to reconstruct more realistic facial details by utilizing information from spatially remote but similar patches and structural facial information. RC-Net \cite{Leng2022RCNetRC} progressively enhances face super-resolution and landmark estimation performance through recurrent collaboration. FSRCH \cite{Lu2022RethinkingPF} improves face reconstruction by extracting facial prior information from original high-resolution images and embedding it into low-resolution ones, introducing high-frequency details. Shi et al. \cite{Shi2022IDPTID} propose IDPT, which employs a pyramid encoder/decoder Transformer architecture to enrich the connection between shallow-layer coarse features and deep-layer detailed features, achieving more effective face super-resolution. Despite these advances, inaccurate facial structure information can still result in significant facial distortions.

The proposed SHT offers superior modeling of facial structure information in two ways: 1) it introduces a unified DHLN (Dual-Head Landmark Network) within a two-stream framework, where the landmark heatmap hallucination stream is enhanced by the face hallucination stream; and 2) unlike other methods \cite{Cao2017AttentionAwareFH, Chen2018FSRNetEL, Ma2020DeepFS} that treat FLD or face hallucination tasks as final supervisory signals, SHT uses these tasks as intermediate objectives. This intermediate supervision effectively enhances the training process, leading to more accurate FLD and face hallucination outcomes.

\section{Supervision-by-Hallucination-and-Transfer}
In this section, we first introduce the proposed DHLN, followed by a detailed description of the FPTN. Finally, we present the overall SHT framework. By integrating the DHLN and FPTN, the facial landmark detection task can be better collaborated with face hallucination and facial pose transfer tasks for achieving precise low-resolution FLD.

\subsection{Dual Hallucination Learning Network}
Current deep learning-based FLD methods \cite{Wu2018LookAB, Dong2018StyleAN, Zhu2019RobustFL, Liu2019SemanticAF, Chandran2020AttentionDrivenCF, Yang2017StackedHN} are still suffering from low-resolution inputs, limiting their effectiveness in practical applications. Inspired by face hallucination \cite{Cao2017AttentionAwareFH, Chen2018FSRNetEL, Ma2020DeepFS} which can maintain high-resolution representations with low-resolution inputs, we propose a Dual Hallucination Learning Network (DHLN as shown in Fig. \ref{fig3}). DHLN aims to address low-resolution facial landmark detection by deeply incorporating the face hallucination task. Specifically, the face hallucination stream can learn high-resolution representations from low-resolution inputs, which helps supplement and enhance the facial pose representations to achieve more precise low-resolution FLD. Simultaneously, the facial structure information learned from the landmark heatmap hallucination stream can be regarded as pose attention and applied to improve face hallucination.

\textbf{Face hallucination stream.} In the proposed DHLN, the face hallucination stream first utilizes the super-resolution block (\textbf{SRBlock} in Fig. \ref{fig3}) to learn high-resolution representations. These high-resolution representations are then fused with the facial pose representations (i.e., the facial structure and shape information learned from the HGBlock) using the proposed fusion block (\textbf{FUBlock1} in Fig. \ref{fig3}). To be specific, the learned facial pose representations are transformed into pose attention, guiding the super-resolution block to focus on key areas of interest, thereby producing more effective pose-attentional high-resolution representations. Suppose $P_t$ denotes the output of the $t$-th Hourglass Network block and $P \in {\mathbb{R}^{64 \times 64 \times {256}}}$, $Q_t$ represents the output of the $t$-th super-resolution module. Each super-resolution module contains 4 super-resolution blocks and $Q \in {\mathbb{R}^{64 \times 64 \times {64}}}$. Then $P_t$ is transformed into $P_t^{'}$ that has the same size as $Q_t$ with a residual block \cite{He2016DeepRL}. After that, $P_t^{'}$ is activated by an element-wise sigmoid function and then multiplied by $Q_t$ to obtain pose attentional high-resolution representations. The whole process can be formulated as follows:

\begin{small}
\begin{equation}
	\begin{array}{l}
		{{Q'}_t} = {Q_t} + {Q_t} \odot \sigma \left( {{P_t}^\prime } \right)\\
		= {Q_t} + {Q_t} \odot \sigma\left( {RB\left( {{P_t}} \right)} \right)
	\end{array}
\end{equation}\end{small}where $\odot$ denotes element-wise product, $RB$ means a residual block, $\sigma$ denotes the sigmoid function and ${{Q'}_t}$ represents the pose attentional high-resolution representations. With such a well-designed fusion block (\textbf{FUBlock1}), the facial structure information learned from the landmark heatmap stream can be utilized to learn the high-resolution information and further help hallucinate more realistic face images.

\textbf{Landmark heatmap hallucination. }Deep learning is commonly used for regression or classification tasks, where the output for an input image is a real number or a single class label. However, in heatmap regression-based FLD, the desired output includes localization, meaning each pixel must be assigned a real number representing its probability. Hence, the low-resolution inputs easily lead to coarse landmark heatmaps and lower detection accuracy. The proposed DHLN addresses this issue by learning high-resolution representations from low-resolution inputs through the integration of facial landmark detection and face hallucination tasks in a two-stream framework. To be specific, the proposed DHLN uses a four Stacked Hourglass Network (SHN) \cite{Yang2017StackedHN} as the backbone of the landmark heatmap hallucination stream to extract facial pose representations. In each Hourglass Network unit, the pose attentional high-resolution representations from the super-resolution module are introduced and fused with the original facial pose representations with a fusion block (\textbf{FUBlock2} in Fig. \ref{fig3}) for obtaining high-resolution facial pose representations. These enhanced representations are then updated and propagated through the proposed DHLN for achieving more precise low-resolution FLD. The entire process can be expressed as follows:

\begin{small}
\begin{equation}
	{P'_t} = {P_t} + conv\left( {{P_t}\left\| {{Q_t}^\prime } \right.} \right)
\end{equation}
\end{small}where $||$ denotes the channel concatenation operation and ${{Q_t}^\prime }$denotes the pose attentional high-resolution representations. $P_t$ and $P_t^{'}$ represent the original and updated high-resolution facial pose representations, respectively. The high-resolution facial pose representations are able to help generate high-resolution landmark heatmaps and achieve more precise low-resolution landmark detection. 

\begin{figure}[t]
\begin{center}
	\includegraphics[width=0.90\linewidth]{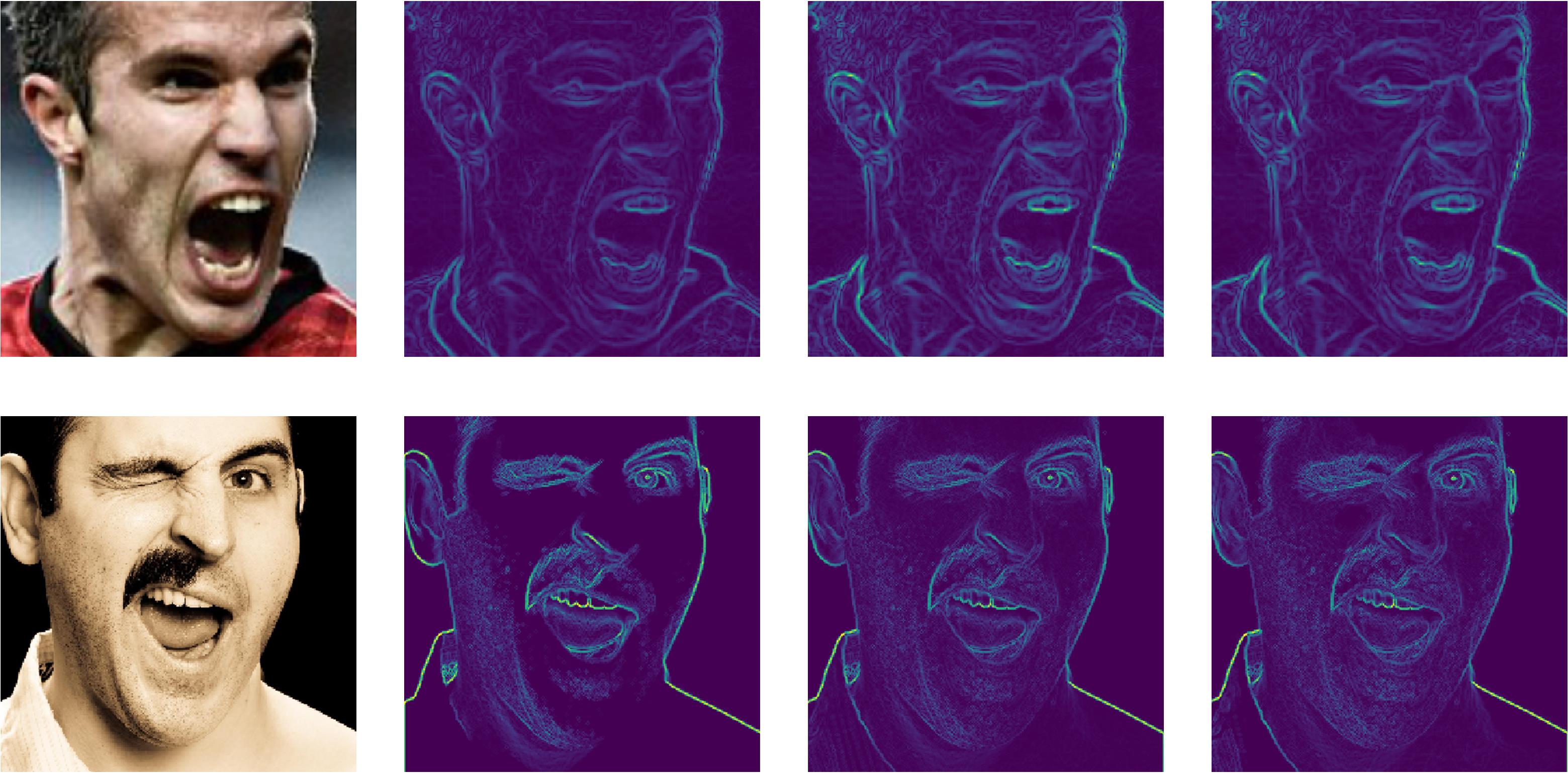}
\end{center}
\caption{Gradient maps. The gradient maps ( corresponding to 3 (RGB) channels) contain rich facial structure information which helps hallucinate more realistic face images.}
\label{fig4}
\end{figure}

\textbf{Objective Function.} The objective function consists of two parts: one for the FLD task and the other for the face hallucination task. For FLD, we utilize the mean squared error loss ($L_2$ loss) between hallucinated landmark heatmaps and the ground-truths. For face hallucination, we use the $L_1$ loss between hallucinated images and the ground-truths as part of the objective function. However, optimizing only the $L_1$ loss tends to produce overly smooth images, which lack sharp edges and clear facial structures. Hence, we introduce the gradient loss \cite{ma2020structure} to enhance facial structure constraints. The gradient loss is computed as the $L_1$ loss between the gradient maps (see Fig. \ref{fig4}) of the hallucinated images and the ground-truths, in which the gradient maps are computed separately for each channel of the image by using convolution layers with fixed kernels. Since most areas of the gradient map are close to zero, except for those corresponding to the facial structure, the gradient loss helps the network focus on the spatial relationships of the outlines. This, in turn, enhances and refines high-resolution facial pose representations, leading to more effective facial landmark detection (FLD) and face hallucination.

Given the training sets $\left\{ {I_i^{LR},I_i^{HR},H_i^*} \right\}_{i = 1}^N$, the following objective function needs to be minimized:

\begin{small}
\begin{equation}
	\begin{array}{l}
		{I_x}\left( {\rm{x}} \right) = I\left( {x + 1,y} \right) - I\left( {x - 1,y} \right),\\
		{I_y}\left( {\rm{x}} \right) = I\left( {x,y + 1} \right) - I\left( {x,y - 1} \right),\\
		\nabla I\left( {\rm{x}} \right) = \sqrt {{{\left( {{I_x}\left( {\rm{x}} \right)} \right)}^2} + {{\left( {{I_y}\left( {\rm{x}} \right)} \right)}^2}} ,\\
		G\left( I \right) = \nabla I,
	\end{array}
\end{equation}\end{small}
\begin{small}
\begin{equation}
	\begin{array}{l}
		{L_{{\rm{DH}}}} = \sum\limits_{i = 1}^N {}  \{ \gamma_1 \left\| {{H_i} - H_i^ * } \right\|_2^2\\
		+ \gamma_2 {\left\| {I_i^{SR} - I_i^{HR}} \right\|_1} + \gamma_3 {\left\| {G\left( {I_i^{SR}} \right) - G(I_i^{HR})} \right\|_1}\} 
	\end{array} 
\end{equation}\end{small}where $i$ denotes the image index and  ${\rm{x}} = (x,y)$. $H$ and $H^*$ denote the generated landmark heatmaps and the ground-truth ones, respectively. $I^{LR}$, $I^{SR}$ and $I^{HR}$ represent the input low-resolution image, hallucinated image and corresponding ground-truth image, respectively. $G$ denotes the gradient loss. By optimizing Eq. (4), FLD and face hallucination tasks can be boosted by each other.

\subsection{Facial Pose Transfer Network}
\begin{figure}[t]
\begin{center}
	\includegraphics[width=0.90\linewidth]{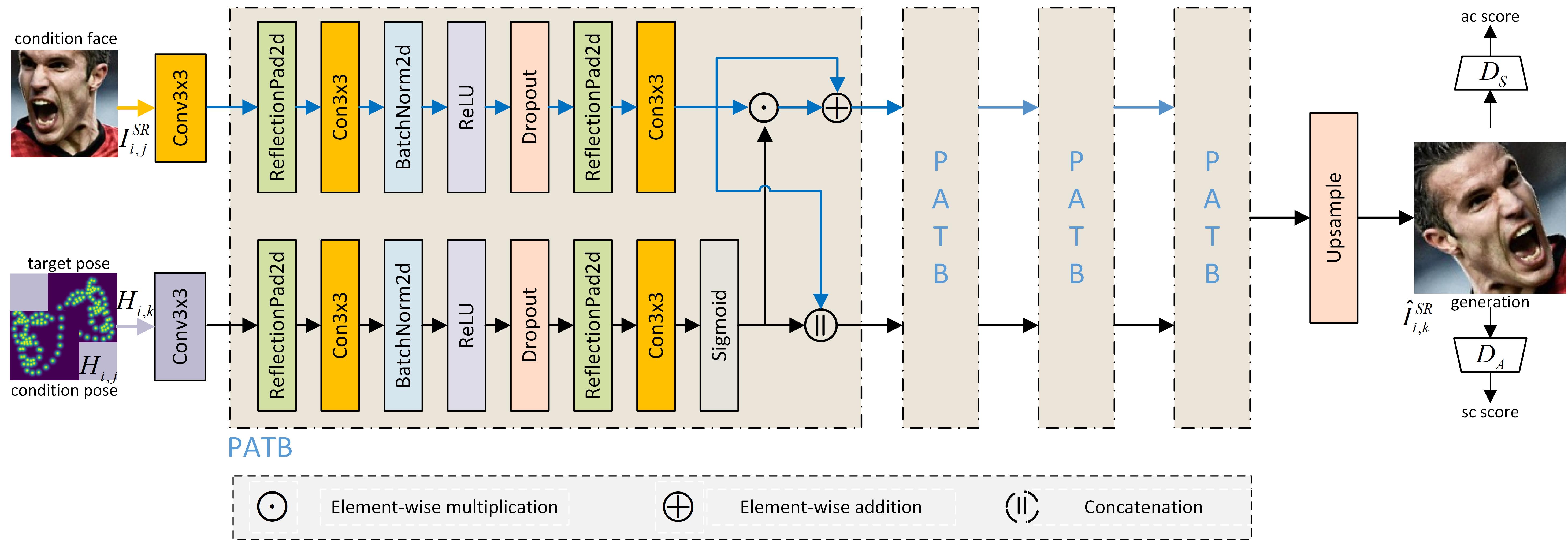}
\end{center}
\caption{The main structure of FPTN. By transforming a person's face from one pose to another, FPTN can improve landmark heatmaps and faces hallucinated by DHLN for achieving more effective FLD and face hallucination.}
\label{fig5}
\end{figure}

The original pose transfer networks \cite{Liu2018PoseTP, Pumarola2018UnsupervisedPI, Zhu2019ProgressivePA} are designed for human pose transfer tasks, while in this paper, we design a new Facial Pose Transfer Network (FPTN) to address the facial pose transfer problem, i.e., transforming a face from one pose to another. Suppose $I_{con}$ and $H_{con}$ denote the condition face image and pose information, and $I_{tar}$ and $H_{tar}$ denote the corresponding target ones. By taking $\left\{ {{I_{con}},{H_{con}},{H_{tar}}} \right\}$ as inputs, the FPTN aims to generate a face image ${I_{ger}}$ that looks the same as ${I_{tar}}$, by optimizing the loss between ${I_{ger}}$ and ${I_{tar}}$. As shown in Fig. \ref{fig5}, we use the landmark heatmaps as the pose information and the \textbf{PATB} \cite{Zhu2019ProgressivePA} as our backbone unit for face generation. By moving patches from conditional pose-induced locations to corresponding target ones, FPTN can generate more realistic face images. The objective function is defined as follows:

\begin{small}
\begin{equation}
	\begin{array}{l}
		{\mathcal{L}_{{\rm{PT}}}}\left( {{I_{con}},{H_{con}},{H_{tar}}} \right)\\
		= \arg \mathop {\min }\limits_G \mathop {\max }\limits_D {\lambda _1}{\mathcal{L}_{GAN}}\left( {{I_{con}},{H_{con}},{H_{tar}},{I_{tar}},{I_{ger}}} \right)\\
		+ {\lambda _2}{\mathcal{L}_{L1}}\left( {{I_{tar}},{I_{ger}}} \right) + {\lambda _3}{\mathcal{L}_{perL1}}\left( {{I_{tar}},{I_{ger}}} \right)
	\end{array}
\end{equation}\end{small}
\begin{small}
\begin{equation}
	\begin{array}{l}
		{\mathcal{L}_{GAN}}\left( {{I_{con}},{H_{con}},{H_{tar}},{I_{tar}},{I_{ger}}} \right)\\
		= {\mathbb{E}_{\scriptstyle{H_{tar}} \in {p_H},\hfill\atop
				\scriptstyle\left( {{I_{con}},{I_{tar}}} \right) \in {p_I}\hfill}}\left\{ {\log [{D_A}\left( {{I_{con}}||{I_{tar}}} \right) \cdot {D_S}\left( {{H_{tar}}||{I_{tar}}} \right)]} \right\}\\
		+ {\mathbb{E}_{\scriptstyle{H_{tar}} \in {p_H},{I_{con}} \in {p_I},\hfill\atop
				\scriptstyle{I_{ger}} \in {{\hat p}_I}\hfill}}\{ \log [\left( {1 - {D_A}\left( {{I_{con}}||{I_{ger}}} \right)} \right)\\
		\cdot \left( {1 - {D_S}\left( {{H_{tar}}||{I_{ger}}} \right)} \right)]\} 
	\end{array}
\end{equation}\end{small}
\begin{small}
\begin{equation}
	{\mathcal{L}_{L1}}\left( {{I_{tar}},{I_{ger}}} \right) = {\left\| {{I_{ger}} - {I_{tar}}} \right\|_1}
\end{equation}\end{small}
\begin{small}
\begin{equation}
	{\mathcal{L}_{perL1}}\left( {{I_{tar}},{I_{ger}}} \right) = {\left\| {\phi \left( {{I_{ger}}} \right) - \phi \left( {{I_{tar}}} \right)} \right\|_1}
\end{equation}\end{small}where $\mathcal{L}_{GAN}$ denotes the adversarial loss, $\mathcal{L}_{L1}$ represents the $L_1$ loss computed between the transferred faces and the ground-truths, and $\mathcal{L}_{perL1}$ means a perceptual $L_1$ loss computed by the VGG-19 model \cite{Simonyan2015VeryDC}. $\phi$ denotes the output of the $Conv1\_2$ layer from the VGG-19 model, $||$ represents the channel concatenation operation. $p_H$, $p_I$ and $\hat p_I$ denote the distribution of real face images, fake face images, and facial poses, respectively. Following \cite{Zhu2019ProgressivePA}, the proposed FTPN also uses two discriminators to keep both appearance and shape consistent. By combining the above three losses, our generated face images look more similar to the input ones in both appearance and shape.
\subsection{Supervision-by-Hallucination-and-Transfer}
The face hallucination stream in DHLN provides high-resolution information to the landmark heatmap hallucination stream, enabling more accurate low-resolution FLD. However, DHLN's performance is still constrained by the availability of training samples. Inspired by how FPTN generates a target face using the condition face, condition landmark heatmaps, and target landmark heatmaps as inputs, we leverage the faces and landmark heatmaps hallucinated by DHLN (i.e., DHLN’s outputs) as inputs for FPTN to further refine and update them. Additionally, since both the condition and target landmark heatmaps can be derived from DHLN's outputs, no manual landmark annotations are required. This allows FPTN to update DHLN using unlabeled high-resolution face images or videos. Therefore, we propose a novel Supervision-by-Hallucination-and-Transfer (SHT) framework (see Fig. \ref{fig2}), which seamlessly integrates FPTN and DHLN to enhance low-resolution FLD and face hallucination in a weakly-supervised way.

Take the simplest case into consideration, a pair of face images $({I_{i,j}^{LR}, I_{i,k}^{LR}})$ is firstly input into DHLN to hallucinate corresponding landmark heatmaps $({{H_{i,j}},{H_{i,k}}})$ and face images $({I_{i,j}^{SR},I_{i,k}^{SR}})$. Here, $i$ denotes the  $i$-th image or the $i$-th video in datasets. For image datasets, $\left( {{I_{i,j}},{I_{i,k}}} \right)$ denote faces with different rotation angels $j$ and $k$ of the $i$-th image, and for video datasets, $\left( {{I_{i,j}},{I_{i,k}}} \right)$ denote faces with different frame indexes $j$ and $k$ of the $i$-th video. Then, $({I_{i,j}^{SR},{H_{i,j}},{H_{i,k}}})$ and $({I_{i,k}^{SR},{H_{i,k}},{H_{i,j}}})$ are input into FPTN to generate corresponding transferred faces $\hat I_{i,k}^{SR}$ and $\hat I_{i,j}^{SR}$, respectively. By optimizing the loss of FPTN, the foreground region of the landmark heatmap hallucinated by DHLN can be further forced to move to the correct position and the faces hallucinated by DHLN can be improved and updated at the same time. Hence, our proposed SHT outperforms state-of-the-art FLD and face hallucination methods. 

The objective function of weakly-supervised SHT is formed as follows:

\begin{small}
\begin{equation}
	\begin{array}{l}
		{\mathcal{L}_{SHT}} = \sum\limits_{n = 1}^N {} \{ {\mathcal{L}_{DH}} + {\mathcal{L}_{{\rm{PT}}}}\left( {{I_{n,j}},{H_{n,j}},{H_{n,k}}} \right)\\
		+ {\mathcal{L}_{{\rm{PT}}}}\left( {{I_{n,k}},{H_{n,k}},{H_{n,j}}} \right)\} \\
		+ \sum\limits_{m = 1}^M {} \{ {{\mathcal{L}'}_{DH}} + {\mathcal{L}_{{\rm{PT}}}}\left( {{I_{m,j}},{H_{m,j}},{H_{m,k}}} \right)\\
		+ {\mathcal{L}_{{\rm{PT}}}}\left( {{I_{m,k}},{H_{m,k}},{H_{m,j}}} \right)\} 
	\end{array}
\end{equation}\end{small}
\begin{small}
\begin{equation}
	\begin{array}{l}
		{{\mathcal{L}'}_{{\rm{DH}}}} = {\rm{\{ }}\sum\limits_m {} (\gamma_2 {\left\| {I_m^{SR} - I_m^{HR}} \right\|_1}\\
		+ {\gamma_3 \left\| {G\left( {I_m^{SR}} \right) - G(I_m^{HR})} \right\|_1}{\rm{\} }}
	\end{array}
\end{equation}\end{small}where $N$ and $M$ denote numbers of the labeled and unlabeled training face images, respectively. ${{\mathcal{L}'}_{{\rm{DH}}}}$ denotes the new DHLN loss that does not include the loss of landmark heatmaps. With Eq. (9) and (10), our SHT can further boost both low-resolution FLD and face hallucination tasks by using unlabeled high-resolution face images.

\section{Experiments}
In this section, we compare our SHT with state-of-the-art face hallucination and FLD methods. Following that, the ablation studies, experimental observations and corresponding analyses are given.
\begin{table*}
\centering
\caption{PSNR, SSIM and $\rm NME_{wid}$ Comparisons with state-of-the-art face hallucination methods. FHS denotes the experimental results of single face hallucination stream, and SHT-v means that we further use 300VW dataset to boost our proposed SHT. }
\begin{center}
	\begin{tabular}{p{3.3cm}|p{0.8cm}p{0.8cm}p{1.3cm}|p{0.8cm}p{0.8cm}p{1.3cm}}
		\hline
		\multirow{2}{*}{Datasets} & \multicolumn{3}{|c|}{CelebA} & \multicolumn{3}{c}{Helen} \\
		\cline{2-7}
		& PSNR$\uparrow$ & SSIM$\uparrow$ &$\rm NME_{wid}$$\downarrow$ & PSNR$\uparrow$ & SSIM$\uparrow$ &$\rm NME_{wid}$$\downarrow$\\
		\cline{2-7}	
		\cline{0-0}
		Bicubic & 23.58 & 0.6285 &0.3385 & 23.89 & 0.6751 &0.4577 \\	
		SRResNet{$\rm _{CVPR17}$}\cite{Ledig2017PhotoRealisticSI} &25.82 &0.7369 &- &25.30 &0.7297 &-   \\
		URDGN{$\rm _{ECCV16}$}\cite{Yu2016UltraResolvingFI}  &24.63 &0.6851 &- &24.22 &0.6909 &- \\
		RDN{$\rm _{CVPR18}$}\cite{Zhang2018ResidualDN}  &26.13 &0.7412 &0.1415 &25.34 &0.7249 &0.4437\\
		PFSR{$\rm _{BMVC19}$}\cite{Kim2019ProgressiveFS} &24.43 &0.6991 &0.1917 &24.73 &0.7323 &0.3498 \\
		FSRNet{$\rm _{CVPR18}$}\cite{Chen2018FSRNetEL}  &26.48 &0.7718 &0.1430 &25.90 &0.7759 &0.3723 \\	
		FSRGAN{$\rm _{CVPR18}$}\cite{Chen2018FSRNetEL}  &25.06 &0.7311 &0.1463 &24.99 &0.7424 &0.3408\\
		DIC{$\rm _{CVPR21}$}\cite{Ma2020DeepFS} &27.37 &0.7962 &0.1320 &26.69 &0.7933 &0.3674 \\
		DICGAN{$\rm _{CVPR21}$}\cite{Ma2020DeepFS} &26.34 &0.7562 &0.1319 &25.96 &0.7624  &0.3336\\   
		AD-GNN{$\rm _{TIP22}$}\cite{Bao2022AttentionDrivenGN} &27.40 &0.7989 &0.1462 &27.71 &0.8306 &0.3693 \\
		IDPT{$\rm _{IJCAI22}$}\cite{Shi2022IDPTID} &27.96 &0.8355 &- &27.59 &0.8045 &- \\
		IDPTGAN{$\rm _{IJCAI22}$}\cite{Shi2022IDPTID} &26.57 &0.7837 &- &26.19 &0.7480  &-\\   	
		FSRCH{$\rm _{TNNLS23}$}\cite{Lu2022RethinkingPF} &27.65 &0.7946 &- &- &- &- \\
		\hline
		\textbf{FHS} & \textbf{25.67} & \textbf{0.7223} &\textbf{-} & \textbf{25.48} & \textbf{0.7352} &\textbf{-} \\
		\textbf{DHLN} & \textbf{27.84} & \textbf{0.8153} &\textbf{0.1279} & \textbf{27.07} & \textbf{0.8177} &\textbf{0.3218} \\
		\textbf{SHT} & \textbf{28.14} & \textbf{0.8238} &\textbf{0.1246} & \textbf{27.48} & \textbf{0.8233} &\textbf{0.3171}  \\ 
		\textbf{SHT-v} & \textbf{28.79} & \textbf{0.8377} &\textbf{0.1198} & \textbf{27.95} & \textbf{0.8310} &\textbf{0.3102} \\ \hline
	\end{tabular}
\end{center}
\label{tabfh}
\end{table*}
\subsection{Datasets and Implementation Details}
\textbf{Datasets.} The facial landmark detection task is evaluated on five datasets including CelebA \cite{Liu2015DeepLF}, Helen \cite{Le2012InteractiveFF}, 300W \cite{Sagonas2016300FI}, AFLW \cite{Zhu2016UnconstrainedFA} and WFLW \cite{Wu2018LookAB}. For the first two datasets, we further use 300VW \cite{Shen2015TheFF} dataset to boost the proposed SHT in a weakly-supervised way. For the last three datasets, we separately use 300VW  and CelebA datasets to train the weakly-supervised SHT.

\textbf{Evaluation Metrics.} For CelebA \cite{Liu2015DeepLF} and Helen \cite{Le2012InteractiveFF} datasets, we use both facial landmark detection and face hallucination evaluation metrics. For 300W \cite{Sagonas2016300FI}, AFLW \cite{Zhu2016UnconstrainedFA} and WFLW \cite{Wu2018LookAB} datasets, we evaluate the proposed SHT with only FLD evaluation metrics, which includes the NRMSE \cite{Bulat2017HowFA, Sun2019HighResolutionRF, Sagonas2016300FI, Ma2020DeepFS}, the AUC \cite{Kumar2020LUVLiFA, Wan2021RobustFA} and the FR \cite{Chen2019FaceAW, Tang2020TowardsEU}. Face hallucination evaluation metrics consist of PSNR and SSIM \cite{Wang2004ImageQA}. For NRMSE, $\rm NME_{box}$ \cite{Bulat2017HowFA, Zafeiriou2017TheMF}, $\rm NME_{diag}$ \cite{Sun2019HighResolutionRF, Wu2018LookAB}, $\rm NME_{io}$ \cite{Deng2019JointMF, Tang2020TowardsEU} and $\rm NME_{wid}$ \cite{Chen2018FSRNetEL, Ma2020DeepFS} are separately utilized to evaluate the FLD performance. AUC is computed as the area under the cumulative error distribution (CED) curve \cite{Kumar2020LUVLiFA, Wan2021RobustFA}. FR refers to the percentage of images in the test set whose $\rm{NME}$ is larger than a certain threshold. The PSNR and SSIM are computed on the $Y$ channel of the transformed YCbCr space.

\textbf{Implementation Details.} The paired images used to train our SHT are generated as follows: for a single face image, we randomly sample the angle of rotation ($-30^\circ, +30^\circ$) and the bounding box scale ($0.8, 1.2$) from a standard Gaussian distribution, and for videos, two frames are randomly selected. During training, the batch size is 16, i.e., 8 pairs of face images. $(\gamma_1, \gamma_2, \gamma_3)$ and $(\lambda_1, \lambda_2, \lambda_3)$ are set to ((0,1), 0.01, 0.01) and (0.05, 0.01, 0.01), respectively. $\gamma_1=1$ corresponds to labeled training samples and  $\gamma_1=0$ for unlabeled ones. SHT is trained with Pytorch on 1 NVIDIA GeForce RTX 4090 GPU. To enhance the performance of our SHT and accelerate its convergence, we employ the following training strategies. For DHLN, we initially train it using a face alignment dataset, such as 300W, with the corresponding loss function defined in Eq. (4). To further improve DHLN, we incorporate FPTN into the model. FPTN is first trained using ground-truth landmark heatmaps to capture facial structure information. After training both models separately, we construct SHT by combining the pretrained DHLN and FPTN and then finetune the combined model using the face alignment dataset. For the weakly-supervised SHT, we extend the finetuning process by utilizing face images or videos without landmark annotations. Specifically, we finetune SHT separately on datasets like CelebA and 300VW, which contain unlabeled data. This allows SHT to learn from a broader range of facial variations.

The proposed SHT aims to improve the FLD and face hallucination performance by integrating DHLN and FPTN, we show the detailed comparisons in the following parts.
\begin{figure*}[t]
\begin{center}
	\includegraphics[width=0.86\linewidth]{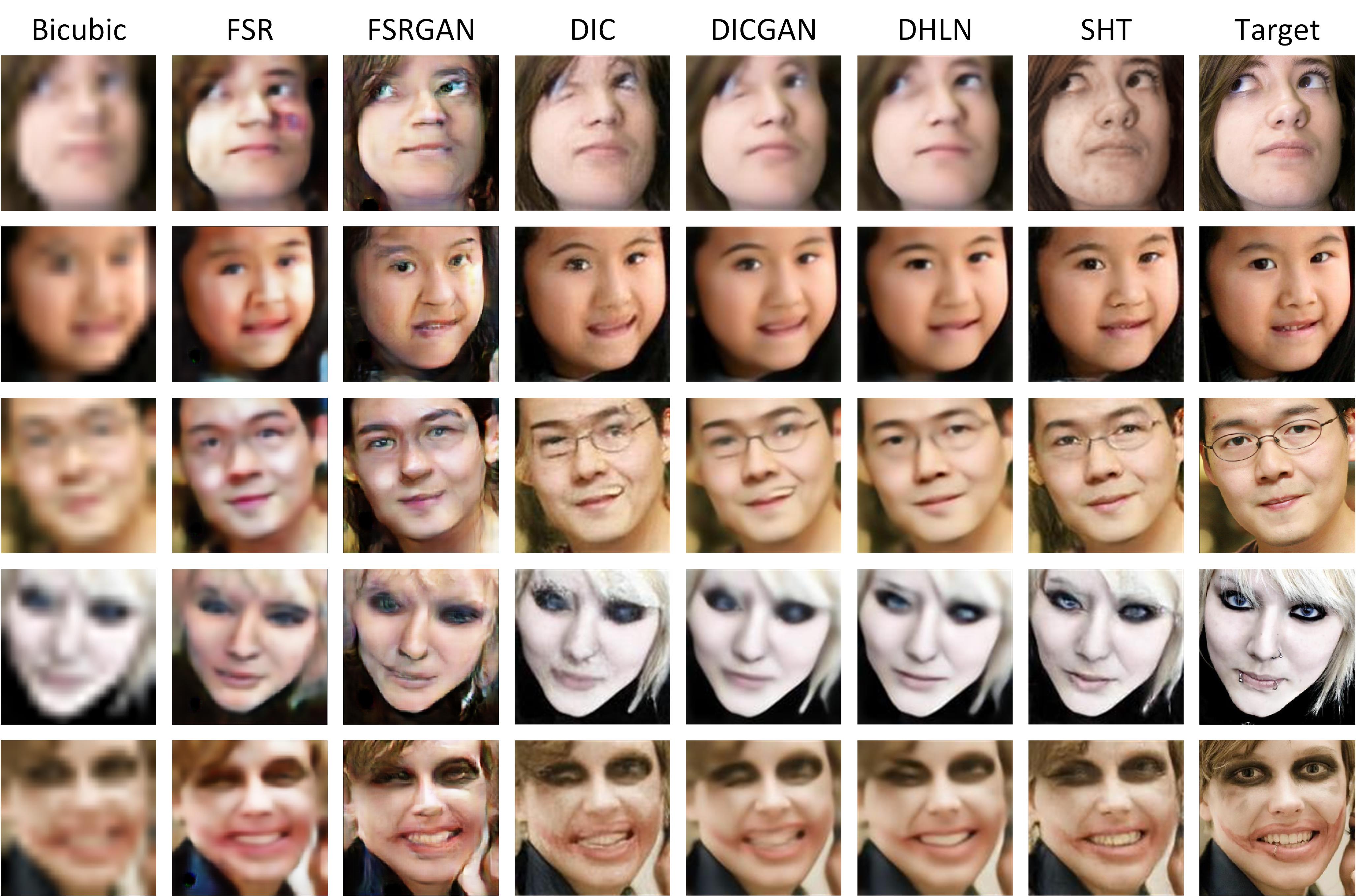}
\end{center}
\caption{Comparison with state-of-the-art face hallucination methods. Our DHLN and SHT perform better in preserving facial structures and generating local details (e.g., face contour, eyes, mouth, paints and eyeglasses) even for faces with large pose and rotation.}
\label{figsr}
\end{figure*}
\subsection{Comparisons with SOTA Methods on CelebA and Helen Datasets}
For CelebA \cite{Liu2015DeepLF} and Helen \cite{Le2012InteractiveFF} datasets, we use both facial landmark detection and face hallucination evaluation metrics to evaluate the proposed method. The detailed experiment settings and result comparisons are shown below.

\textbf{Experiment Settings.} SRResNet \cite{Ledig2017PhotoRealisticSI}, URDGN \cite{Yu2016UltraResolvingFI}, RDN \cite{Zhang2018ResidualDN}, FSRNet \cite{Chen2018FSRNetEL}, DIC \cite{Ma2020DeepFS}, IDPT \cite{Shi2022IDPTID} and FSRCH \cite{Lu2022RethinkingPF},  and our proposed method all address face hallucination problems by incorporating facial landmark detection task. They usually take $16 \times 16$ (LR) face images as inputs and output $128 \times 128$ (HR) ones. For fair comparisons, we first obtain $16 \times 16$ face images with bicubic degradation, then resize them to $64 \times 64$ and input them into DHLN. Then, we combine DHLN and FPTN to construct \textbf{SHT} method and further boost SHT with an additional 300VW dataset (denoted by \textbf{SHT-v}).

\textbf{CelebA. }The training set consists of 168,854 images, while the testing set includes 1,000 images. The detected 68 landmarks from OpenFace \cite{Baltruaitis2018OpenFace2F} are treated as ground-truth annotations. As shown in Table \ref{tabfh}, our proposed SHT achieves the highest PSNR score and the second-highest SSIM score among the evaluated methods \cite{Bao2022AttentionDrivenGN, Shi2022IDPTID, Lu2022RethinkingPF}. Furthermore, landmark detection accuracy is enhanced by utilizing the unlabeled 300VW dataset. Simultaneously, the face hallucination performance improves progressively as we integrate SHT, demonstrating that the face hallucination stream and landmark heatmap stream effectively complement and reinforce each other.

\textbf{Helen.} The training set contains 2,005 images, and the testing set includes 50 images. Table \ref{tabfh} presents a comparison of landmark detection and face hallucination results on the Helen dataset \cite{Le2012InteractiveFF} against state-of-the-art methods \cite{Ma2020DeepFS, Bao2022AttentionDrivenGN, Shi2022IDPTID, Lu2022RethinkingPF}. We can find that our SHT-v achieves the best $\rm NME_{wid}$, PSNR and SSIM scores. Visual comparisons of face hallucination results in Fig. \ref{figsr} show that our DHLN and SHT outperform other methods, recovering more accurate details and producing more complete facial structures. These suggest that by deeply cooperating with landmark heatmap hallucination stream and facial pose transfer network, our SHT preserves pixel-wise accuracy while increasing the perceptual quality of the hallucinated images and achieving more precise low-resolution landmark detection. Additionally, both face hallucination and facial landmark detection are further improved by introducing more facial prior knowledge from unlabeled face videos.

\begin{table}
\caption{$\rm NME_{io}$ comparisions on 300W dataset. SHN is the baseline and M means ``modified''. SHT-M-i and SHT-M-v represent we separately use CelebA and 300VW datasets to boost our proposed SHT-M.}
\footnotesize
\begin{center}
	\begin{tabular}{p{4.8cm}|p{1.5cm}p{2cm}p{1.5cm}}
		\hline
		Method  & 
		\begin{tabular}[c]{@{}c@{}}Common\\ Subset\end{tabular}$\downarrow$ & \begin{tabular}[c]{@{}c@{}}Challenging\\ Subset\end{tabular}$\downarrow$ & Fullset $\downarrow$ \\ \hline                                                      
		SAN{$\rm _{CVPR18}$}\cite{Dong2018StyleAN}           & 3.34    & 6.60      & 3.98    \\
		AVS{$\rm _{ICCV19}$}\cite{Qian2019AggregationVS} & 3.21    & 6.49      & 3.86    \\
		LAB{$\rm _{CVPR18}$}\cite{Wu2018LookAB}            & 2.98    & 5.19      & 3.49    \\
		Techer{$\rm _{ICCV19}$}\cite{Dong2019TeacherSS}  & 2.91    & 5.91      & 3.49    \\
		DU-Net{$\rm _{ECCV18}$}\cite{Tang2018QuantizedDC}    & 2.90    & 5.15      & 3.35    \\
		DeCaFa{$\rm _{ICCV19}$}\cite{Dapogny2019DeCaFADC} & 2.93    & 5.26      & 3.39    \\
		HR-Net{$\rm _{19'}$}\cite{Sun2019HighResolutionRF}               & 2.87    & 5.15      & 3.32    \\
		HG-HSLE{$\rm _{ICCV19}$}\cite{Zou2019LearningRF} & 2.85    & 5.03      & 3.28    \\
		AWing{$\rm _{ICCV19}$}\cite{Wang2019AdaptiveWL}     & 2.72  & 4.52   & 3.07 \\
		
		LUVLi{$\rm _{CVPR20}$}\cite{Kumar2020LUVLiFA}       & 2.76  & 5.16    & 3.23 \\  
		ADNet{$\rm _{ICCV21}$}\cite{huang2021adnet}       & 2.53  &4.58    & 2.93
		\\ 
		GlomFce{$\rm _{CVPR22}$}\cite{Zhu2022OcclusionrobustFA}       & 2.72  &4.48    & 3.13
		\\
		SLPT{$\rm _{CVPR22}$}\cite{Xia2022SparseLP}       & 2.75  &4.90    & 3.17
		\\
		DTLD+{$\rm _{CVPR22}$}\cite{Li2022TowardsAF}       & 2.60  &4.48    & 2.96		
		\\
		STAR+{$\rm _{CVPR23}$}\cite{Zhou2023STARLR}       & 2.52  &4.32    & 2.87		
		\\
		\hline
		SHN ($256 \times 256$)  & 3.11  & 6.23    & 3.72  \\ 
		\textbf{DHLN-M} & \textbf{2.74}  & \textbf{4.78}    & \textbf{3.14}  \\ 
		\textbf{SHT-M} & \textbf{2.57}  & \textbf{4.23}    & \textbf{2.90}  \\ 
		\textbf{SHT-M-i} & \textbf{2.46}  & \textbf{4.07}    & \textbf{2.78}  \\ 
		\textbf{SHT-M-v} & \textbf{2.50} & \textbf{4.14}    & \textbf{2.82}  \\ \hline
	\end{tabular}
\end{center}
\label{tab300w}
\end{table}
\subsection{Comparisons with SOTA Methods on the other Datasets}
For the other datasets including 300W \cite{Sagonas2016300FI}, AFLW \cite{Zhu2016UnconstrainedFA} and WFLW \cite{Wu2018LookAB}, we use only facial landmark detection evaluation metrics to evaluate our proposed methods. The corresponding experiment settings and result comparisons are shown below.

\indent\textbf{Experiment Settings.} Current mainstream FLD methods \cite{Wu2018LookAB, Liu2019SemanticAF, Wang2019AdaptiveWL, Wan2021RobustFA, huang2021adnet} mainly use $256 \times 256$ (HR) images as inputs. Therefore, we should modify our SHT for fair comparisons, which is denoted as \textbf{SHT-M}. SHT-M inputs $64 \times 64$ (LR) images, but hallucinates $256 \times 256$ (HR) ones. By optimizing Eq. (4), the original high-resolution information can also be used in our proposed SHT-M. To construct the new SHT-M, we should add an upsample layer to the face hallucination stream before outputting the hallucinated faces. Then, FPTN should also make the corresponding modification. The $64 \times 64$ (LR) images are obtained from the $256 \times 256$ (HR) images with bicubic degradation. The detailed experiments are conducted as follows. As the main backbone of the landmark heatmap hallucination stream is the Hourglass Network unit, we use the SHN ($256 \times 256$) \cite{Yang2017StackedHN} as the baseline. DHLN-M is constructed by combining landmark heatmap hallucination stream and face hallucination stream. After that, we also construct SHT-M by integrating DHLN-M and FPTN-M. Moreover, as SHT-M can work in a weakly-supervised way, we also use CelebA \cite{Liu2015DeepLF} and 300VW \cite{Shen2015TheFF} to train SHT-M, represented by \textbf{SHT-M-i} and \textbf{SHT-M-v} respectively. For CelebA, the bounding boxes are obtained according to the detected 68 landmarks with OpenFace \cite{Baltruaitis2018OpenFace2F}. Next, we will show detailed comparisons with state-of-the-art FLD methods on challenging datasets.

\textbf{300W.} The whole dataset is further divided into four subsets: Training set (3148 images), Common subset (554 images), Challenging subset (135 images), and Fullset (689 images). From Table \ref{tab300w}, we find SHT-M beats other FLD methods \cite{Kumar2018Disentangling3P, Dong2018StyleAN, Qian2019AggregationVS, Wu2018LookAB, Kumar2020LUVLiFA} in both Challenging Subset and Fullset \cite{Kumar2020LUVLiFA}. The reason is that the test set contains many low-resolution faces, while current FLD methods directly resize those faces to high-resolution ones and then take them as inputs, which reduces the detection accuracy. Our method can address this problem because it could learn high-resolution representations from low-resolution inputs through deep collaboration with the face hallucination stream. 

\textbf{AFLW.} AFLW \cite{Zhu2016UnconstrainedFA} contains 19 landmarks, with many face images featuring large poses and significant occlusions. The training set consists of 20,000 images, and the testing set is divided into two subsets: AFLW-full (4,386 images) and AFLW-frontal (1,314 images). As presented in Table \ref{tabaflw}, our SHT surpasses state-of-the-art methods \cite{Robinson2019LaplaceLL, Dong2018StyleAN, Miao2018DirectSR, Wu2018LookAB} across the $\rm NME_{diag}$, $\rm NME_{box}$, and $\rm AUC^7_{box}$ metrics. These results highlight that by integrating the Dual Hallucination Learning Network (DHLN) and the Facial Pose Transfer Network (FPTN) into the novel SHT framework, the landmark heatmap generated by DHLN-M is further refined, leading to more accurate facial landmark detection (FLD).

\begin{table}
\caption{$\rm NME$ and $\rm AUC$ comparisions on AFLW dataset. M means ``modified'', SHT-M-i and SHT-M-v represent we separately use CelebA and 300VW datasets to boost our proposed SHT-M.}
\begin{center}
	\begin{tabular}{p{4.cm}|p{1.2cm}p{1.3cm}|p{1.5cm}|p{1.5cm}}
		\hline
		\multirow{2}{*}{Method} & \multicolumn{2}{|c|}{$\rm NME_{diag}$$\downarrow$} &{$\rm NME_{box}$$\downarrow$} & {$\rm AUC^7_{box}$$\uparrow$} \\
		\cline{2-5}
		& Full & Frontal & Full & Full \\
		\cline{2-5}	
		\cline{0-0}
		LLL{$\rm _{ICCV19}$}\cite{Robinson2019LaplaceLL} & 1.97 & - & - & -  \\
		SAN{$\rm _{CVPR18}$}\cite{Dong2018StyleAN} & 1.91 & 1.85 & - & -  \\
		DSRN{$\rm _{CVPR18}$}\cite{Miao2018DirectSR} & 1.86 & - & - & -  \\	
		LAB{$\rm _{CVPR18}$}\cite{Wu2018LookAB} & 1.85 & 1.62 & - & -  \\
		HR-Net{$\rm _{19'}$}\cite{Sun2019HighResolutionRF} & 1.57 & 1.46 & - & -  \\
		Wing{$\rm _{CVPR18}$}\cite{Feng2018WingLF} & - & - & 3.56 & 5.35  \\ 
		KDN\cite{Chen2018KernelDN} & - & - & 2.80 & 60.3  \\ 
		MMDN{$\rm _{TNNLS22}$}\cite{Wan2020RobustFL} & 1.41 & 1.21 & - & -  \\
		LUVLi{$\rm _{CVPR20}$}\cite{Kumar2020LUVLiFA} & 1.39 & 1.19 & 2.28 & 68.0  \\ 
		MHHN{$\rm _{TIP21}$}\cite{Wan2021RobustFA} & 1.38 & 1.19 & - & -  \\ 
		DTLD{$\rm _{CVPR22}$}\cite{Li2022TowardsAF} & 1.37 & - & - & -  \\
		SCPAN{$\rm _{TCYB23}$} \cite{Wan2021RobustAP} &1.31 &1.10 &2.05 &69.8 \\
		semi-SCPAN{$\rm _{TCYB23}$} \cite{Wan2021RobustAP} &1.23 &1.05 &2.01 &70.7 \\
		\hline
		\textbf{SHT-M} & \textbf{1.21} & \textbf{1.06} & \textbf{2.14} & \textbf{70.1}  \\ 
		\textbf{SHT-M-i} & \textbf{1.09} & \textbf{0.96} & \textbf{2.01} & \textbf{72.4}  \\ 
		\textbf{SHT-M-v} & \textbf{1.13} & \textbf{0.99} & \textbf{2.05} & \textbf{71.6}  \\ \hline
	\end{tabular}
\end{center}
\label{tabaflw}
\end{table}
\begin{table}
\caption{$\rm NME_{io}$, $\rm AUC^{10}_{io}$ and $\rm FR^{10}_{io}$ comparisions on WFLW dataset. M means ``modified'', SHT-M-i and SHT-M-v represent we separately use CelebA and 300VW datasets to boost our proposed SHT-M.}
\begin{center}	
	\begin{tabular}{p{4.2cm}|p{2cm}|p{2cm}|p{2cm}}
		\hline
		Method & $\rm NME_{io}$$\downarrow$  &$\rm AUC^{10}_{io}$$\uparrow$ & $\rm FR^{10}_{io}$$\downarrow$ \\ \hline
		CFSS{$\rm _{CVPR15}$}\cite{Zhu2015FaceAB} & 9.02 & 0366 & 20.56  \\	
		DVLN{$\rm _{CVPR17}$}\cite{Wu2017LeveragingIA} & 10.84 & 0.456 & 10.84   \\	
		LAB{$\rm _{CVPR18}$}\cite{Wu2018LookAB} & 5.27 & 0.532 & 7.56 \\
		Wing{$\rm _{CVPR18}$}\cite{Feng2018WingLF} & 5.11 & 0.554 & 6.00   \\ 
		DeCaFa{$\rm _{ICCV19}$}\cite{Dapogny2019DeCaFADC} & 4.62 & 0.563 & 4.84  \\ 
		AVS{$\rm _{ICCV19}$}\cite{Qian2019AggregationVS} & 4.39 & 0.591 & 4.08 \\
		AWing{$\rm _{ICCV19}$}\cite{Wang2019AdaptiveWL} & 4.36 & 0.572 & 2.84\\
		LUVLi{$\rm _{CVPR20}$}\cite{Kumar2020LUVLiFA} & 4.37 & 0.577 & 3.12  \\
		ADNet{$\rm _{ICCV21}$}\cite{huang2021adnet} & 4.14  &0.602    & 2.72  \\
		SLPT{$\rm _{CVPR22}$}\cite{Xia2022SparseLP} & 4.14  &0.595    & 2.76  \\
		DTLD{$\rm _{CVPR22}$}\cite{Li2022TowardsAF} & 4.08  &-    & 2.76  \\
		DTLD+{$\rm _{CVPR22}$}\cite{Li2022TowardsAF} & 4.05  &-    & 2.68  \\
		STAR{$\rm _{CVPR23}$}\cite{Zhou2023STARLR} &4.02 &0.605  &2.32\\ \hline
		\textbf{SHT-M} & \textbf{4.03} & \textbf{0.605} & \textbf{2.48 }  \\ 
		\textbf{SHT-M-i} & \textbf{3.92} & \textbf{0.621} & \textbf{2.36}  \\ 
		\textbf{SHT-M-v} & \textbf{3.96} & \textbf{0.618} & \textbf{2.44}  \\ \hline
	\end{tabular}
\end{center}
\label{tabwflw}
\end{table}

\textbf{WFLW.} The WFLW dataset \cite{Wu2018LookAB} contains 98 landmarks, with 7,500 images used for training and 2,500 images for testing. From the comparisons in Table \ref{tabwflw} on $\rm NME_{io}$, $\rm AUC^{10}{io}$, and $\rm FR^{10}{io}$ metrics, our method outperforms other approaches \cite{Zhu2015FaceAB, Wu2017LeveragingIA, Wu2018LookAB, Qian2019AggregationVS, Wang2019AdaptiveWL, Kumar2020LUVLiFA}. These results demonstrate that SHT-M is more robust to various facial variations, including occlusion, pose, make-up, blur, and expression.

\begin{table*}
\caption{Impact of gradient loss on PSNR, SSIM and $\rm NME_{wid}$ metrics. [* = without gradient loss]}
\begin{center}
	\begin{tabular}{p{1.8cm}|p{1.2cm}p{1.2cm}p{1.3cm}|p{1.2cm}p{1.2cm}p{1.2cm}}
		\hline
		\multirow{2}{*}{Datasets} & \multicolumn{3}{|c|}{CelebA} & \multicolumn{3}{c}{Helen} \\
		\cline{2-7}
		& PSNR$\uparrow$ & SSIM$\uparrow$ &$\rm NME_{wid}$$\downarrow$ & PSNR$\uparrow$ & SSIM$\uparrow$ &$\rm NME_{wid}$$\downarrow$\\
		\cline{2-7}	
		\cline{0-0}
		FHS & 25.67 &0.7223 &- &25.48 &0.7352 &- \\\hline
		DHLN* &27.79 & 0.8010 &0.1285 &27.01 &0.8161 &0.3231 \\  
		DHLN &27.84 & 0.8153 &0.1279 &27.07 &0.8177 &0.3218 \\  
		SHT* &28.07 &0.8223 &0.1254 &27.34 &0.8213 &0.3183  \\
		SHT &28.14 &0.8238 &0.1246 &27.48 &0.8233 &0.3171  \\
		SHT-v* & 28.70 &0.8366 &0.1207 &27.89 &0.8301 &0.3109 \\
		SHT-v &28.79 &0.8377 &0.1198 &27.95 &0.8310 &0.3102 \\ \hline
	\end{tabular}
\end{center}
\label{gradient}
\end{table*}

\textbf{Weakly-supervised SHT.} The proposed SHT can be further boosted by utilizing face images or videos without landmark annotations in a weakly-supervised way. Therefore, we separately combine CelebA and 300VW datasets to train the proposed SHT. 169854 images in CelebA \cite{Liu2015DeepLF} and 50 training videos (95192 frames) in 300VW \cite{Shen2015TheFF} datasets are separately used for training. For example, we train SHT-M-i on 300W by combining 300W training set and CelebA, in which the ratio of labeled data (3148) to the whole data (173002) is 1.8\%. From Tables \ref{tab300w}, \ref{tabaflw} and \ref{tabwflw}, we can see that both SHT-M-i and SHT-M-v outperform the original SHT-M, which indicates that SHT-M can be further enhanced with unlabeled face data. This improvement can be attributed to the incorporation of additional facial prior knowledge, allowing SHT-M-i and SHT-M-v to generate more realistic face images and more accurate landmark heatmaps. Moreover, we also compare our SHT with other weakly-supervised methods \cite{Dong2018, Dong2019, Wan2021RobustAP}. For a fair comparison, we conduct experiments by combining YouTube-Face \cite{Wolf2011FaceRI} and AFLW datasets, which are respectively denoted as SHT-M (+YouTube-Face) and SHT-M (+AFLW). The results, presented in Table \ref{weakly}, show that SHT-M (+YouTube-Face) surpasses SBR (+YouTube-Face), demonstrating the efficacy of our weakly-supervised approach. However, the performance of both SHT-M (+AFLW) and SHT-M-i (+CelebA) is slightly lower than that of TS (+AFLW) and Semi-SCPAN (+CelebA). This slight discrepancy may be due to the fact that our SHT-M model uses lower-resolution images (i.e., $64 \times 64 \times 3$) as inputs.

\begin{table*}
	\caption{${\rm NME_{ip}}$ Comparsion on 300W dataset with weakly-supervised methods.}
	\begin{center}
		\begin{tabular}{p{4.8cm}|p{1.5cm}p{2cm}p{1.5cm}}
			\hline
			Method  & 
			\begin{tabular}[c]{@{}c@{}}Common\\ Subset\end{tabular}$\downarrow$ & \begin{tabular}[c]{@{}c@{}}Challenging\\ Subset\end{tabular}$\downarrow$ & Fullset $\downarrow$ \\ \hline                                                      
			SBR(+YouTube-Face) \cite{Dong2018}	&3.28	&7.58&	4.10  \\ 
			\textbf{SHT-M(+YouTube-Face)}	&2.98	&6.07	&3.59\\ 
			TS(+AFLW) \cite{Dong2019}	&2.91	&5.91	&3.49\\ 
			\textbf{SHT-M(+AFLW)} &	2.94 &	5.97&	3.53\\ 
			Semi-SCPAN(+CelebA) \cite{Wan2021RobustAP}	&2.92&	5.96&	3.52\\ 
		\textbf{SHT-M-i(+CelebA)}	&	2.97&	6.12&	3.59 \\ \hline
		\end{tabular}
	\end{center}
	\label{weakly}
\end{table*}

\begin{figure}[t]
\begin{center}
	\includegraphics[width=0.96\linewidth]{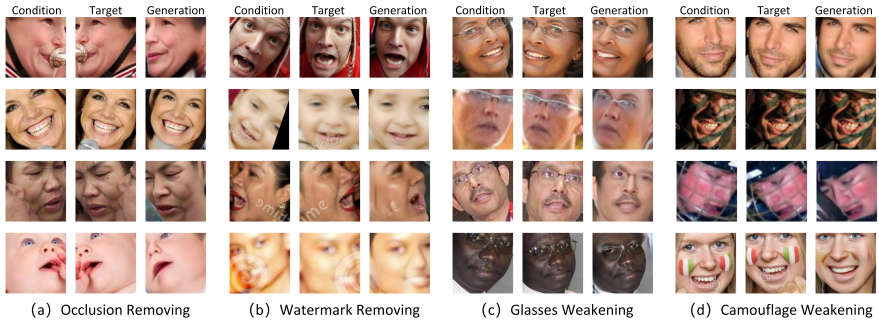}
\end{center}
\caption{Visualization of transferred faces of SHT-M-i on WFLW dataset. With rich prior knowledge, our SHT-M-i can perform effective face restoration and keep/recover good facial structures, thus helping boost both facial landmark detection and face hallucination. }
\label{fig7}
\end{figure}

\textbf{Face Transfer. }By transferring faces and utilizing unlabeled high-resolution face images or videos, our proposed SHT is able to generate more realistic faces. We combine CelebA \cite{Liu2015DeepLF} to train our proposed SHT-M-i and the transferred faces are shown in Fig. \ref{fig7}. ``Condition" and ``Target" refer to the ground-truth high-resolution images, and ``Generation" represents the transferred face. We can see that 1) our SHT-M-i can remove occlusions (e.g., mouthpiece, microphone, face and hand as shown in Fig. \ref{fig7} (a)), and watermask (Fig. \ref{fig7} (b)) for faces no matter the condition one or the target one is occluded or not, and 2)  our SHT-M-i can weak glasses (Fig. \ref{fig7} (c)), beard or camouflage ( (Fig. \ref{fig7} (d))) for faces under these covers. Therefore, we can conclude that by integrating DHLN and FPTN via the proposed SHT framework, more facial prior knowledge can be introduced from unlabeled face data to simultaneously boost both face hallucination and facial landmark detection.

\subsection{Ablation Studies}
We conduct the ablation studies from the following three aspects.

\textbf{Gradient Loss.} Since the gradient map helps the network focus on the spatial relationships of facial outlines, it also purifies and enhances high-resolution facial pose representations. To validate this, we conducted ablation studies comparing performance with and without the gradient loss. As shown in Table \ref{gradient}, introducing the gradient loss improves both facial landmark detection (FLD) and face hallucination, demonstrating its effectiveness in enhancing the overall performance of the network.

\textbf{Face Hallucination.} As the face hallucination stream can also hallucinate high-resolution faces by stacking SRBlocks, we also statistic the performance of the single face hallucination stream (denoted as \textbf{FHS}) for comparisons. As shown in Table \ref{tabfh}, our SHT-v obtains the best PSNR, SSIM and $\rm NME_{wid}$ scores, our DHLN and SHT also outperform the other methods in both face hallucination and FLD metrics. Moreover, SHT and SHT-v both surpass DHLN, and DHLN outperforms FHS. These experimental results indicate that 1) landmark heatmap hallucination stream and face hallucination stream can be boosted by each other via a two-stream framework; 2) by integrating DHLN and FPTN, our SHT can recover more correct local details and complete facial structure and 3) the additional facial prior knowledge from unlabeled face datasets can be introduced by our proposed SHT framework to further boost face hallucination.

\textbf{Facial Landmark Detection.} The backbone of our landmark heatmap hallucination stream is a four stacked Hourglass network (i.e., SHN \cite{Yang2017StackedHN}), so we use SHN $256 \times 256$ as the baseline. Then, we combine the landmark heatmap hallucination stream and face hallucination stream to design our dual hallucination learning network (DHLN-M). As shown in Table \ref{tab300w}, Our DHLN-M can achieve 2.74\%, 4.78\% and 3.14\% $\rm{NME_{io}}$ on 300W Common subset, 300W Challenging subset and 300W full set, which outperforms the SHN $256 \times 256$ (3.11\% , 6.23\% and 3.72\%). When we further integrate the DHLN-M and the FPTN-M with the proposed SHT framework, the accuracy of FLD can be improved. From Table \ref{tab300w}, we can see that SHT-M exceeds both DHLN-M and other methods. More importantly, when SHT-M is trained with additionally unlabeled face images or videos, the detection accuracy (see SHT-M-i and SHT-M-v in Tables \ref{tab300w}, \ref{tabaflw}, and \ref{tabwflw}.) are further boosted. These results indicate that 1) landmark heatmap hallucination stream can be boosted by incorporating face hallucination stream; 2) the performance of FLD can be further improved by integrating DHLN-M and FPTN-M via the proposed SHT-M and 3) SHT-M can utilize unlabeled face images or videos to achieve more effective FLD in a weakly-supervised framework.

\subsection{Self Evaluations}
\textbf{Time and Memory Analysis. } SHT combines DHLN and FPTN, where FPTN can be freely removed during the inference phase and thus does not incur additional overhead. DHLN consists of two streams: the face hallucination stream and the landmark hallucination stream. The landmark hallucination stream in DHLN shares a similar network structure with SHN \cite{Yang2017StackedHN}, which contains 16.41M parameters. Meanwhile, the face hallucination stream introduces multiple SRBlocks for learning high-resolution features, as well as FUBlock1 and FUBlock2, which facilitate the interaction between facial pose and high-resolution representations. These additional components increase the parameter count slightly, bringing DHLN to 22.72M parameters, slightly more than SHN. Despite this increase, the added parameters are negligible relative to the 24 GB memory of an RTX 4090 GPU, and the computational cost will continue to diminish with the advancement of hardware technologies. But on 300W common subset, our DHLN can achieve a 17.36\% performance improvement over SHN. In terms of model size, DHLN occupies 228 MB, while SHN takes up 184 MB. Additionally, DHLN can achieve a reference speed of 80 FPS on a single RTX 4090 GPU, compared to SHN's 120 FPS. 

\subsection{Experimental Results and Discussions}
According to the above experimental results, we have the following observations and corresponding analyses.

(1) SHN \cite{Yang2017StackedHN}, SAN \cite{Dong2018StyleAN}, HR-net \cite{Sun2019HighResolutionRF}, AWing \cite{Wang2019AdaptiveWL}, HG-HSLE \cite{Zou2019LearningRF}, LUVLi \cite{Kumar2020LUVLiFA} and our SHT-M are all belong to heatmap regression-based FLD methods. However, our SHT-M surpasses the other methods as shown in Tables \ref{tab300w}--\ref{tabwflw}. This indicates that 1) by integrating landmark heatmap hallucination stream and face hallucination stream via the proposed DHLN in a two-stream framework, high-resolution deep representations can be learned from low-resolution inputs to achieve more precise FLD; 2) the performance of both FLD and face hallucination can be further boosted by incorporating the facial pose transfer network and 3) by utilizing unlabeled high-resolution face images or videos, more facial prior knowledge can be introduced to address FLD for faces with variations on head poses and occlusions.

(2) SRResNet \cite{Ledig2017PhotoRealisticSI}, URDGN \cite{Yu2016UltraResolvingFI}, RDN \cite{Zhang2018ResidualDN}, FSRNet \cite{Chen2018FSRNetEL}, DIC \cite{Ma2020DeepFS} and our proposed method all address facial landmark detection problems by incorporating face hallucination task. While our proposed DHLN and SHT surpass the other methods, which indicates that 1) the landmark heatmap hallucination stream in the proposed DHLN can provide more effective facial structural information that can help generate more realistic faces without distortion and 2) by integrating the FPTN, faces generated by DHLN can be further updated and improved for preserving pixel-wise accuracy and increasing the perceptual quality.

(3) As shown in Tables \ref{tab300w}, \ref{tabaflw} and \ref{tabwflw}, SHT-M-i beats SHT-M-v, which mainly because 1) the CelebA \cite{Liu2015DeepLF} dataset contains more peoples than 300VW \cite{Shen2015TheFF} and the transformation operations (random rotation and scaling) on CelebA will also produce large-scale paired images that can provide rich facial prior knowledge and 2) it is more difficult for FPTN to transfer face image poses with rotations outside of the image plane than in the image plane.

(4) As shown in the first two rows of Fig. \ref{fig7}, when we combine 300VW or CelebA to train SHT, the occlusion and blurring problems of the transferred face images can be reduced, which indicates that our proposed SHT can address face de-occlusion and de-blurring problems by introducing additional facial prior information, thereby achieving more precise and robust FLD.
\section{Conclusion}
Precise and robust FLD remains a great challenge due to low-quality input information and insufficient labeled training samples. In this work, a novel SHT framework is proposed to address this challenge in a weakly-supervised way. It is shown that the two streams in DHLN can be boosted by each other for hallucinating faces without distortions and meanwhile generating more effective landmark heatmaps. Moreover, by incorporating FPTN, SHT can reduce the dependence on labeled datasets and enhance the performance of both FLD and face hallucination. Experimental results on FLD and face hallucination demonstrate that our SHT outperforms state-of-the-art methods. It can also be found from the experiment that by combining face hallucination and facial pose transfer tasks, our SHT could also use rich facial prior knowledge to address the problem of face de-occlusion and de-blurring, which further improves the detection accuracy. 

\leftline{ {\bf Acknowledgements}} This work is supported by the National Natural Science Foundation of China (Grant No. 62571555), the Natural Science Foundation of Hubei Province, China (Grant No. 2024AFB992).

\section*{References}

\bibliography{mybibfile}

\end{document}